\documentclass{article}

     \usepackage[preprint, nonatbib]{neurips_2019}

\usepackage{wrapfig}
\usepackage[utf8]{inputenc} 
\usepackage[T1]{fontenc}    
\usepackage{hyperref}       
\usepackage{url}            
\usepackage{booktabs}       
\usepackage{amsfonts}       
\usepackage{nicefrac}       
\usepackage{microtype}      
\usepackage{booktabs} 
\usepackage{diagbox}
\usepackage{multirow}
\usepackage{booktabs,array,xcolor,colortbl,multicol,epsfig,graphicx,subfigure}
\usepackage{amsmath,amsthm,amsfonts,amssymb,bm}
\usepackage[vlined,boxed,commentsnumbered,ruled,linesnumbered]{algorithm2e}
\usepackage{caption}
\usepackage{colortbl,accents}
\usepackage{subfigure}
\usepackage{caption}
\usepackage{color}
\def\widebar{\accentset{{\cc@style\underline{\mskip10mu}}}}
\def\Widebar{\accentset{{\cc@style\underline{\mskip8mu}}}}
\makeatother

\title{Reinforcement Learning with Policy Mixture Model\\ for Temporal Point Processes Clustering}

\author{%
	Weichang Wu \\
	Shanghai Jiao Tong University\\
	\texttt{blade091@sjtu.edu.cn} \\
	\And
	Junchi Yan \thanks{Corresponding author.}\\
	Shanghai Jiao Tong University \\
	\texttt{yanjunchi@sjtu.edu.cn} \\
	\AND
	Xiaokang Yang \\
	Shanghai Jiao Tong University \\
	\texttt{xkyang@sjtu.edu.cn} \\
	\And
	Hongyuan Zha \\
	Georgia Institute of Technology \\
	\texttt{zha@cc.gatech.edu} \\
}

\begin{document}

\maketitle

\begin{abstract}
	Temporal point process is an expressive tool for modeling event sequences over time. In this paper, we take a reinforcement learning view whereby the observed sequences are assumed to be generated from a mixture of latent policies. The purpose is to cluster the sequences with different temporal patterns into the underlying policies while learning each of the policy model. The flexibility of our model lies in: i) all the components are networks including the policy network for modeling the intensity function of temporal point process; ii) to handle varying-length event sequences, we resort to inverse reinforcement learning by decomposing the observed sequence into states (RNN hidden embedding of history) and actions (time interval to next event) in order to learn the reward function, thus achieving better performance or increasing efficiency compared to existing methods using rewards over the entire sequence such as log-likelihood or Wasserstein distance. We adopt an expectation-maximization framework with the E-step estimating the cluster labels for each sequence, and the M-step aiming to learn the respective policy. Extensive experiments show the efficacy of our method against state-of-the-arts.

\end{abstract}
\vspace{-4pt}
\section{Introduction}
\vspace{-6pt}
Event sequences with time stamps in continuous domain are ubiquitous across different areas and applications. In e-commerce, online purchase records over time can form event sequences. In health informatics, a series of treatments taken by patient can be tracked as an event sequence. In seismology, a sequence of earthquakes can be recorded. 
Recognizing and understanding the structure in the event sequences is of vital importance for downstream applications.

Temporal point processes (TPP)~\cite{daley2007introduction} are useful tools for modeling event sequences whereby one key concept is to model the event occurrence rate over time using the conditional intensity function. Lots of literatures have been proposed for both data-mining like in \cite{zhou2013learning,li2014learning} and prediction task like in \cite{yan2013towards}. 

In this paper, we are devoted to another important but relatively less studied scenario: event sequence clustering in the continuous time domain, which can be more challenging than the traditional (aggregated and discrete) time series clustering. Event sequence clustering can find its utility in many real-world applications. Given a number of event sequences, it is important to discover and learn the underlying clustering structure robustly. For example, the purchase records can help cluster e-commerce users into different groups to benefit a recommender system; clustering patients according to their treatment logs helps hospitals optimize medication resources.


However, despite the extensive existing works on event sequence modeling and prediction as mentioned above, event sequence clustering and especially learning mixture model of event sequences have rarely been addressed, except in \cite{XuNIPS17} to our best knowledge. In \cite{XuNIPS17}, a parametric likelihood based latent Dirichlet allocation model is proposed for event sequences clustering using Hawkes processes. In this paper, we propose a deep reinforcement learning (RL) based EM framework for event sequence clustering using likelihood-free temporal point processes learning, and the purpose is to discover and learn these underlying policies of experts which can be shared over similar sequences. 
he highlights of our work include:

1) We present a network based EM framework for TPP clustering, differing from previous work using parametric clustering models \cite{XuNIPS17}. Under the EM scheme, clustering of the entire dataset and model fitting of each cluster are jointly performed rather than separated in two steps \cite{han2013transition}.

2) We take an RL view to the TPP clustering problem, whereby each cluster corresponds to one of the latent expert policies that generates the observed sequences in the cluster. Our method can be seen as a meta learning approach by adopting IRL to learn policy's reward function. 
In particular, we employ generative adversarial imitation learning \cite{DBLP:conf/nips/HoE16} as an efficient IRL embodiment for TPP.

3) We empirically show that our method exceeds peer models notably including a very recent one \cite{XuNIPS17}. We also compare with the recent mixture GAN based (image) clustering model \cite{DBLP:conf/ijcai/YuZ18} and adapt it to temporal event sequences by using the Wasserstein distance between point processes according to \cite{XiaoNIPS17} (see details in supplementary material), for a more fair comparison. The results show that our model outperforms significantly regarding training efficiency (one order faster), with similar performance.
\vspace{-4pt}
\section{Related Work}
\vspace{-6pt}
We review TPP methods in literature in two aspects: i) modeling of the intensity function; ii) learning objectives and algorithms. The relevance lies in that both the intensity function and learning objective relate to latent policy discovering and learning.


\subsection{Temporal Point Process Intensity Modeling}
Traditional TPP models are mostly developed around the design of the intensity function $\lambda(t)$ which measures the instantaneous event occurrence rate at time $t$, like Reinforced Poisson processes \cite{PemantlePS07}, Self-exciting process (Hawkes process) \cite{hawkes1971point}, Reactive point process \cite{ErtekinRPP2015}, etc. An obvious limitation of these traditional models is that they all assume all the samples obey a single parametric form which is too idealistic for real-world data. This also suggests the need for learning clustered behaviors beyond single model based methods. 


By contrast, in neural point process \cite{DuKDD16,mei2017neural,XiaoAAAI17}, recurrent neural network (RNN) and its variants e.g. long-short term memory (LSTM) are used for modeling the conditional intensity function over time. As such, the learning can be generally fulfilled by gradient descent and no restriction on the form of the intensity function is imposed. More recently attention mechanisms are introduced to improve the interpretability of the neural model \cite{wang2017cascade}. However, in all these models, the event sequences are all fed into the model without any discrimination to the groups they may differently belong to. In fact, the behind mechanisms for generating each event sequence can be very different, thus it may be a better idea to learn one model for each cluster of similar samples for more accurate modeling.

\subsection{Temporal Point Process Learning}
There are alternative objectives for TPP learning for both parametric models and neural models. Traditional methods mostly follow the maximum likelihood estimation (MLE) procedure under the probabilistic framework \cite{ozaki1979maximum}. While the MLE objective may not be the only choice. This is because we are often given only a limited number of sequences which may further contain arbitrary noises. Recent efforts have been made for devising adversarial learning based objective inspired by generative adversarial networks (GAN) \cite{goodfellow2014generative} and especially Wasserstein GAN \cite{ArjovskyWGAN17}. In \cite{YanIJCAI18}, adversarial objective is developed in addition with MLE loss by approximating the continuous domain predictions using a discrete time series. In \cite{XiaoNIPS17}, Wasserstein distance over temporal event sequences is explicitly defined to learn a deep generative point process model for temporal events generation.

Another line of works consider the challenge for learning high-dimensional TPP models, whereby the so-called infectivity matrix to be learned can be of squared size of the dimensionality. One popular technique is imposing low-rank regularizer \cite{ZhouAISTATS13} or factorization model \cite{WuKDD18} on the infectivity matrix. However, they do not explicitly deal with the sequence clustering problem. In fact, the observed dimension marker does not correspond to the underlying cluster.

The mostly related work to our method appears in \cite{XuNIPS17} as it deals with a similar problem setting: grouping event sequences into different clusters and learn the respective TPP model parameters for each cluster. However, the technical approaches are completely different. First the parametric model \cite{XuNIPS17} is tailored to Hawkes process while our network based model is more general; Second, the work \cite{XuNIPS17} is under the Bayesian probabilistic framework while our method is likelihood-free and incorporates both adversarial learning and inverse reinforcement learning~\cite{NgICML00} for more effective objective design beyond MLE; We show in the experiments that our method significantly outperforms \cite{XuNIPS17} on real-world data. Source code will be made public available for reproducible research.

\begin{figure*}[tb!]
	\centering
	\subfigure{\includegraphics[width=0.96\textwidth]{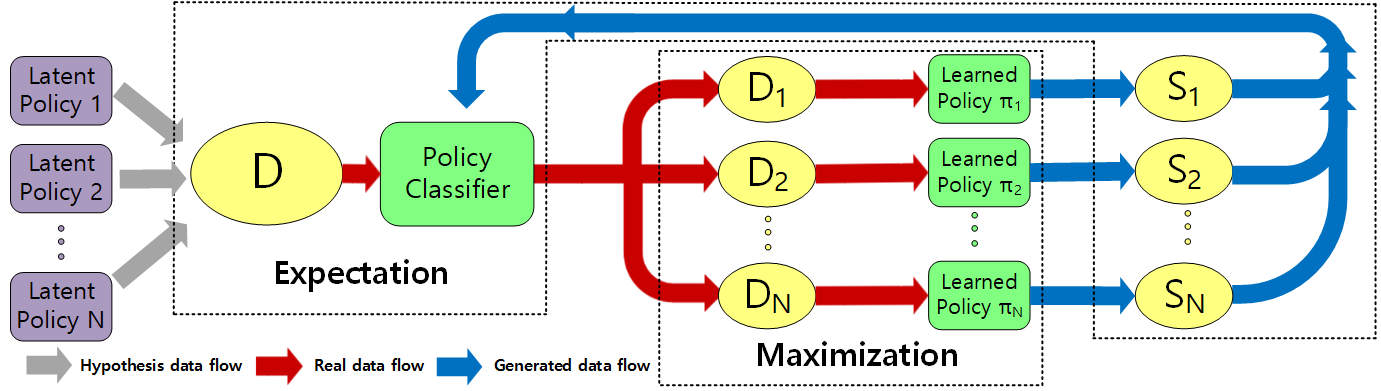}}\vspace{-10pt}
	\caption{EM framework for TPP clustering using deep RL. Dataset $D$ is supposed to be generated by a mixture of $N$ latent policies $\{\pi_n\}_{n=1}^N$. In E-step, fix the learned $\{\pi_n\}_{n=1}^N$ and update the $N$-class classifier $h_q$ by using $N$ sets of event sequences $\{S_n\}_{n=1}^N$ generated from $\pi_n$ with cluster label $n$. In M-step, fix classifier $h_q$ and update each of $\{\pi_n\}_{n=1}^N$ by subset $D_i$ classified by $h_q$.}\label{fig:model}
	\vspace{-8pt}
\end{figure*}

As shown in Fig.~\ref{fig:model}, this paper takes a reinforcement learning (RL) perspective on the modeling and clustering of temporal point processes for its dynamic sequence nature. Though there exist works~\cite{farajtabar2017fake,LiNIPS18,UpadhyayNIPS18} on (deep) RL and intervention of TPP, while little effort (\cite{XuNIPS17} does not involve deep model nor RL) has been paid on TPP clustering which calls additional careful treatment on disentangling the mixture of policies. Using the language of RL, suppose there is a number of event sequences generated (with noise) by $N$ underlying expert policies, which can be reflected in the form of $N$ clusters. In this sense, we formulate the event sequence clustering task as a reinforcement learning problem whereby the purpose is to discover the unknown event generation policies, and meanwhile the learning cost function for fitting event sequences is also automatically learned from data using IRL.%

\vspace{-4pt}
\section{Proposed Model}
\vspace{-6pt}
We present our approach under the expectation-maximization (EM) framework -- a natural way for disentangling the clustering and fitting of each cluster of event sequences which are assumed to be generated by a mixture of policies. In expectation step, each sequence is assigned with a cluster label corresponding to its latent policy that generates this sequence. The latent policy of each cluster is learned in maximization step. The model is illustrated in in Fig. \ref{fig:model}. The complete learning algorithm is shown in Alg. \ref{alg:EM} which calls an IRL based subfunction \emph{GAIL-TPP} described in Alg. \ref{alg:GAIL}.
\begin{algorithm}[tb!]
	\caption{\textbf{Reinforcement Learning for Policy Mixture Model(RLPMM) for TPP }}
	\label{alg:EM}
	\KwIn{dataset $D$, number of policies $N$;\\
	\quad\quad\quad training set size $m=256$ for classifier $h_q$; \\\quad\quad\quad learning rate $\alpha=1e^{-4},\beta=1e^{-4}$; $k=0$.}
	randomly initialize classifier $h_q$, $N$ policies' parameters $\theta_i^{(0)}$, discriminators' parameters $w_i^{(0)}$ for $i=1,2,\dots,N$;
	randomly divide $D$ into $\{D_1^{(0)}, D_2^{(0)},\dots,D_N^{(0)}\}$;
	$\{\theta_i^{(1)}, w_i^{(1)}\}_{i=1}^{N}\leftarrow \emph{GAIL-TPP}(\theta_i^{(0)}, w_i^{(0)}, D_i^{(0)})$;\\
	//\emph{E-step}: L\ref{line:e_sample}, L\ref{line:e_train}, L\ref{line:e_label}; \emph{M-step}: L\ref{line:e_gail}\\
    \While {$\pi_\theta$ not converged}
	{
		sample a training set $S=\{x_j, y_j\}_{j=1}^{m}$ from $\{\pi_{\theta_i^{(k)}}\}_{i=1}^N$ with probability $\frac{|D_i^{(k-1)}|}{|D|}$;\\\label{line:e_sample}
		train classifier $h_q$ using $S$;\\\label{line:e_train}
		compute policy $y_i=h_q(x_i)$ to label $D$ into $\{D_i\}_{i=1}^{N}$;\\\label{line:e_label}
		$\{\theta_i^{(k+1)}, w_i^{(k+1)}\}_{i=1}^{N}\leftarrow \emph{GAIL-TPP}(\theta_i^{(k)}, w_i^{(k)}, D_i^{(k)})$;\\\label{line:e_gail}
		$k=k+1$;
	}
	\KwOut{learned $N$ latent polices $\pi_\theta^* = \{\pi_{\theta_i^{(k)}}\}_{i=1}^N$}
\end{algorithm}
\subsection{EM Learning for Policy Mixture Model}
Given a temporal event set $X$ with $M$ observed event sequences: $X=\{x_1,x_2,\dots,x_M\}$ and the  discrete latents i.e. cluster labels $Y=\{y_1, y_2, \dots, y_M\}$ for $y_i\in\{1,2,\dots,N\}$, we suppose that $X$ are generated by a mixture of $N$ experts with a latent policy for each expert, parameterized by $\theta$ as a whole. The log likelihood is:
$$\mathcal{L}(\theta;X,Y) = \log p(X,Y|\theta),$$
where $p(X,Y|\theta)$ is the conditional probability of observing $X,Y$ given parameter $\theta$, and $\theta$ is determined by maximizing the marginal log likelihood of observed $X$:
$$\theta^* = \mathop{\arg\max}\limits_\theta \mathcal{L}(\theta;X)=\mathop{\arg\max}\limits_\theta \log p(X|\theta),$$
where $p(X|\theta)=\int p(X,Y|\theta)dY$ is the marginal probability.

Suppose the latents $Y$ are sampled from an arbitrary valid probability distribution $q(Y)$, then a lower bound $\mathcal{F}(q, \theta)$ of the marginal log likelihood $\mathcal{L}(\theta;X)$ can be obtained by Jensen's inequality as:
\begin{equation}
\mathcal{F}(q, \theta)=\mathcal{L}(\theta;X) - D_{KL}(q||p),
\end{equation}
which means we have $\forall q(Y):\mathcal{L}(\theta;X)\geq\mathcal{F}(q,\theta)$\footnote{The detailed derivation is presented in supplementary material in the EM learning convergence proof}. Given randomly initialized parameter $\theta^{(0)}$ and arbitrary distribution $q^{(0)}$, we iteratively update $q^{(k)}$ and parameter $\theta^{(k)}$ by the following Expectation-Maximization procedure:

\textbf{E-step}: given model parameter $\theta^{(k)}$, update $q^{(k)}$ to $q^{(k+1)}$ by matching $q$ to posterior $p(Y|X,\theta^{(k)})$

\textbf{M-step}: given $q^{(k+1)}$, update $\theta$: $\theta^{(k+1)} \leftarrow \theta^{(k)} + \nabla \mathcal{F}(q^{(k+1)}, \theta)$ by maximizing $\mathcal{F}(q^{(k+1)}, \theta)$

We iteratively perform E-step and M-step until $\theta^*$, and the distribution of the hidden variable $Y$ is:
$
q^*(Y)=p(Y|X,\theta^*).
$
A detailed proof for the convergence of the EM procedure 
is presented in the supplementary material.

Specifically, the computational components in E-step and M-step are all implemented by neural networks as the Reinforcement Learning with Policy Mixture Model (RLPMM), including the \emph{E-step for policy clustering} and \emph{M-step for policy learning}.

\begin{algorithm}[tb!]
	\caption{\small{\textbf{Generative Adversarial Imitation Learning  \cite{DBLP:conf/nips/HoE16} for TPP: GAIL-TPP ($\theta_i^{(k)},w_i^{(k)},D_i$)}}}
	\label{alg:GAIL}
	\KwIn{set $D_i$, discriminator parameter $w_i^{(k)}$, policy net $\theta_i^{(k)}$}
	sample sequence $x_i \sim \pi_{\theta_i}$\;
	IRL: update discriminator parameters from $w_i^{(k)}$ to $w_i^{(k+1)}$ with gradient computed by Eq.~\ref{GAIL-IRL}\;
	RL: update policy parameters from $\theta_i^{(k)}$ to $\theta_i^{(k+1)}$ with gradient computed by Eq.~\ref{GAIL-RL}\;
	\KwOut{parameter of policy networks $\theta_i^{(k+1)}$, and discriminators $w_i^{(k+1)}$}
\end{algorithm}
\subsection{Expectation Step for Policy Clustering}
As mentioned above, in E-step, we match the hidden variable distribution $q$ to posterior distribution $p(Y|X,\theta^{(k)})$, and fill in values of latent variables $Y$ for samples in observed data $X$ according $q(Y)$, so that we can re-recompute the expectation of $X$ given $\theta^{(k)}$, i.e., the likelihood function $\mathcal{L}(\theta^{(k)})$.

We compute the hidden variable distribution $q$ as
\begin{equation} \label{KL}
	q^{(k)} = \mathop{\arg\min}\limits_{q\in \mathcal{H}_k} KL(q||p^{(k)}),
\end{equation}
where we restrict the distribution of hidden variable $q^{(k)}$ is in a bounded hypothesis space $\mathcal{H}_k$.

For mixture of policies, given parameter $\theta^{(k)}$ and observed data $X$, the posterior distribution $p^{(k)}$ is:
\begin{equation} \label{posterior}
	p(y_{ij}|x_j, \theta^{(k)}) = \frac{p(x_j, y_{ij}|\theta^{(k)})}{p(x_j|\theta^{(k)})},
\end{equation}
where $y_{ij}=1$ if and only if $x_j$ is generated by the $i$-th policy.

Inspired by Eq.~\ref{posterior}, to find $q^{(k)}$ in Eq.~\ref{KL}, we train a classifier to fit the current guess of the discrete hidden variable distribution $q^{(k)}$ to $p^{(k)}$, i.e., holding the policies parameter $\theta^{(k)}$ fixed, train a classifier $h_q$ by data generated by learned policies. Therefore, the E-step involves line \ref{line:e_sample}, \ref{line:e_train}, \ref{line:e_label} in Alg. \ref{alg:EM}.

In practical application, $h_q$ is a $3$-layer classifier including sequence embedding layer, RNN layer and classification layer as used in \cite{DuKDD16,XiaoAAAI17}. In addition, an implementation trick dealing with imbalanced classification at the beginning of the procedure is used, as presented in the supplementary material. 

\subsection{Maximization Step for Policy Learning}

Given the hidden variable $Y$ estimated by the classifier in E-step, each event sequence $x$ in the training dataset $D$ is classified to a specific policy with discrete hidden variable $y_i$. In M-step, dataset $D$ is divided into $N$ clusters $\{D_1, D_2, \dots, D_N\}$ to train each policy model.

Now we present an IRL based policy learning method. Differing from previous works imposing a specific form for sequence learning e.g. Wasserstein distance~\cite{XiaoNIPS17}, we assume the policy reward (cost) is unknown which need be learned via IRL. The rationale is that it is nontrivial to define the temporal event sequence fitting error with varying length in contrast to the vector-like data. To make the learning scalable to real-world data, the IRL procedure is efficiently fulfilled by a generative adversarial imitation learning scheme \cite{DBLP:conf/nips/HoE16}.
\subsubsection{Policy network for sequence generation}\label{subsubsec:policy_net}
\begin{wrapfigure}[17]{r}{0.48\textwidth}
	\vspace{-23pt}
	\includegraphics[width=0.48\textwidth]{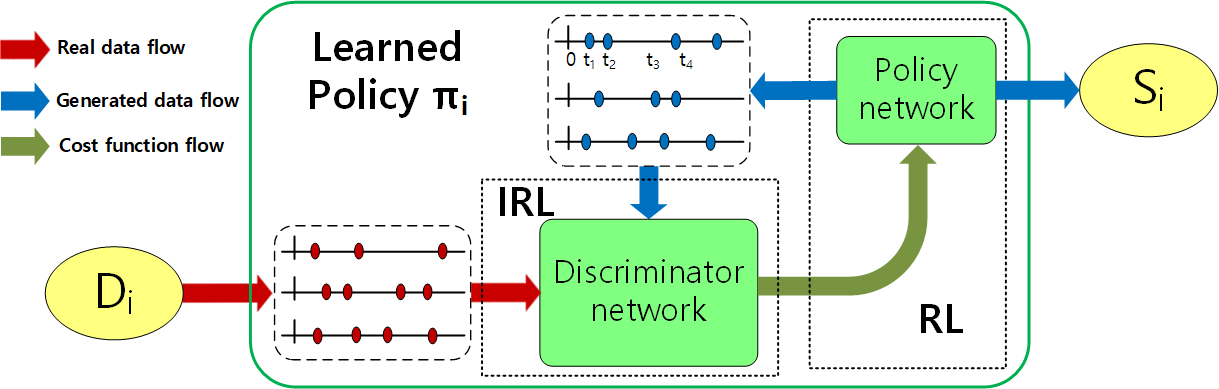}
	\caption{Generative Adversarial Imitation Learning for TPP employed in M-step (see policy $\pi_i$ in green block in Fig.~\ref{fig:GAIL}): Given subset $D_i$ for policy $\pi_i$, for reward (cost) function learning in IRL-step, discriminator is trained by Eq.~\ref{GAIL-IRL} using generated fake sequences and real sequences. In RL-step, policy network is trained by policy gradient using Eq.~\ref{GAIL-RL}, using the learned cost. Back to Fig.~\ref{fig:GAIL}, sequences $\{S_i\}$ generated by policy network are used to train classifier $h_q$ in E-step.}\label{fig:GAIL}
\end{wrapfigure}
The policy function $\pi_\theta$ should have the capacity to capture the complex sequential dependency pattern and stochastic nature in the point process. We adopt RNN with stochastic neurons \cite{bayer2014learning} as the policy network. Here action refers to the time to next event from current event timestamp and state refers to the hidden embedding of RNN for the history.
Note action $a$ is sampled from distribution $\pi(a|\theta(h_{i-1}))$ as
\begin{equation}\label{equ:policy}
	a_i\thicksim \pi(a|\theta(h_{i-1})),
\end{equation}
where $h_i=\psi(Va_i + Wh_{i-1})$, $a_i=t_i-t_{i-1}\in\mathbb{R}^+$ is the $i$-th inter-event time ($t_0=0$), $h_i\in\mathbb{R}^d$ is hidden state of RNN encoding history before $t_i$, $\theta$ is nonlinear mapping from $\mathbb{R}$ to policy's parameter space, $V\in\mathbb{R}^d$ and $W\in\mathbb{R}^{d\times d}$ are coefficients, $\psi$ is nonlinear activation function, e.g., $\psi(z)=\frac{e^{z}-e^{-z}}{e^{z}+e^{-z}}$ as used in this paper. 

There are alternatives for parameterizing the policy function, i.e. the probability density function $\pi$, as long as they satisfy the constraint that the random variable sampled from $\pi$ is positive since $a>0$, such as exponential distribution: $\pi(a|\theta(h_{i-1}))=\theta(h)e^{-\theta(h)a}$ and Rayleigh distribution: $\theta(h)ae^{-\theta(h)a^2/2}$ as used in this paper.

So far the RNN policy network with stochastic neurons is able to mimic the event generating mechanism of stochastic temporal point process by Eq.~\ref{equ:policy}. Given a sequence of past events $s_t=\{t_i\}_{t_i<t}$, the next event time is generated as $t_{i+1} = t_i + a$, with the inter-event time $a$ sampled from stochastic policy $\pi_\theta(a|s_t)$ as the action.

\subsubsection{Inverse reinforcement learning for cost modeling}

As shown in Fig.\ref{fig:GAIL}, we use the GAIL framework to learn the cost function for reinforcement learning, in which the IRL procedure is substituted by training a Discriminator $D_w$, with the gradient of discriminator parameter $w$ given by
\begin{equation}\label{GAIL-IRL}
	\mathbb{E}_{x_\theta}[\nabla_w\log(D_w(s,a))] + \mathbb{E}_{x_E}(\nabla_w\log(1-D_w(s,a)),
\end{equation}
where $x_\theta \sim \pi_\theta$ is the sequences sampled from learned policy $\pi_\theta$ and $x_E \sim \pi_E$ is the sequences sampled from expert's true policy. 

Given the cost function $\log D_w(s,a)$, the RL procedure is implemented by the policy gradient descent with gradient of policy network parameter $\theta$ given by
\begin{equation}\label{GAIL-RL}
	\mathbb{E}_{x_\theta}[\nabla_\theta \log\pi_\theta(a|s)Q(s,a)] - \lambda\nabla_\theta H(\pi_\theta),
\end{equation}
where $Q(\bar{s}, \bar{a}) = \mathbb{E}_{x_\theta}[\log(D_w(s,a))|s_0=\bar{s}, a_0=\bar{a}]$. And the gradient of the  causal entropy regularizer is given by
\begin{align}\label{eq:ent_reg}
	\nabla_\theta H(\theta) &=\nabla_\theta\mathbb{E}_{\pi_\theta}[-\log_{\pi_\theta}(a|s)] \\\notag
	& = \mathbb{E}_{\pi_\theta}[\nabla_\theta \log \pi_\theta (a|s) Q_{\log} (s,a)],
\end{align}
where $Q_{\log}(\bar{s}, \bar{a}) = \mathbb{E}_{\pi_\theta} [-\log_{\pi_\theta}(a|s) | s_0=\bar{s}, a_0=\bar{a}]$

In essence, by iteratively updating $w$ using IRL in Eq.~\ref{GAIL-IRL}, and updating $\theta$ using RL in Eq.~\ref{GAIL-RL}, the GAIL algorithm find a saddle point $(\pi_\theta^*, D_w^*)$ of the expression
\begin{equation}
\mathbb{E}_{\pi_\theta} [\log(D(s,a))] + \mathbb{E}_{\pi_E} [\log(1-D(s,a))] - \lambda H(\pi),
\end{equation}
which is equivalent to find the optimal policy $\pi_\theta^*$.

\begin{table*}[tb!]
	\centering
	\resizebox{0.98\textwidth}{!}{
		\begin{tabular}{r|cccccc}
			\toprule
			Method&Intensity function& Framework& Running pipeline& Learning model \\\midrule
            \textbf{RLPMM}& Neural networks (RNN) & Temporal point process &Joint clustering and learning & Reinforcement learning\\
			WGANMM \cite{DBLP:conf/ijcai/YuZ18}& Neural networks (RNN) & Temporal point process &Joint clustering and learning & Adversarial learning\\
            DMMHP \cite{XuNIPS17}& Parametric Hawkes & Temporal point process &Joint clustering and learning & Maximum likelihood learning\\
            ODE/LS+GMM \cite{LewisJNS2011,eichler2017graphical}& Nonparametric Hawkes  & Temporal point process &Separate clustering and learning & Feature based GMM\\
             VAR+GMM \cite{han2013transition}& Not applicable & Discretized time series &Separate clustering and learning & Feature based GMM\\
			\bottomrule
	\end{tabular}}
	\caption{Comparison including WGANMM from \cite{DBLP:conf/ijcai/YuZ18} which is adapted to TPP data in this paper.}\label{tab:comp_methods}
	\vspace{-15pt}
\end{table*}
\vspace{-4pt}
\section{Experiments}
\vspace{-6pt}
We evaluate the performance of our \textbf{Reinforcement Learning for Policy Mixture Model (RLPMM)} on both synthetic and real-world data. To demonstrate the effectiveness and efficiency of our model, we compare with state-of-the-art methods for event sequence clustering. 

As summarized in Table \ref{tab:comp_methods}, peer methods include: \textbf{1) Gaussian Mixture Model (GMM)} includes 3 Two-step pipeline models that firstly extract features from sequential events using Vector Auto-Regression (VAR)\cite{han2013transition}, Ordinary Differential Equation (ODE)\cite{LewisJNS2011} or Least Squares (LS)\cite{eichler2017graphical}, the use the GMM to cluster the event sequences; \textbf{2) Dirichlet Mixture Model of Hawkes Processes (DMMHP)}\cite{XuNIPS17} which is the most related work to our knowledge; \textbf{3) Wasserstein Generative Adversarial Network Mixture Model (WGANMM)} that we adapt from \cite{DBLP:conf/ijcai/YuZ18}. Due to page limit, we present the technical details of our adaption and the other baselines in the supplementary material.

Metrics used to measure clustering performance are \textbf{1) Clustering Purity (CP)}~\cite{schutze2008introduction}; \textbf{2) Rand Index (RI)}~\cite{rand1971objective}; \textbf{3) Empirical Intensity Deviation (EID)}~\cite{XiaoNIPS17}; \textbf{4) Clustering Consistency (CC)}~\cite{Tibshirani2005Cluster}. All the metrics 
except for clustering consistency (as CC itself already involves random trials) 
are computed by the average of 10 trials on the whole dataset by random initialization for clustering. The details of their definitions are also presented in supplementary material.

\begin{table}[tb!]
	\centering
	\resizebox{\textwidth}{!}{
		\begin{tabular}{c|cccccc|cccccc}
			\toprule
			{Model} & VAR+ & ODE+ & LS+ & \multirow{2}{*}{DMMHP} & {WGAN} & \multirow{2}{*}{\textbf{RLPMM}} & VAR+ & ODE+ & LS+ & \multirow{2}{*}{DMMHP} & {WGAN} & \multirow{2}{*}{\textbf{RLPMM}} \\
			Cluster\#& GMM & GMM & GMM &  &-MM & &  GMM & GMM & GMM &  &-MM& \\ \midrule
			{Evaluation} & \multicolumn{12}{c}{Clustering Purity (CP): the higher the better}  \\\midrule
            $\text{Dataset}$ & \multicolumn{6}{c|}{$\text{noHawkes}$ } & \multicolumn{6}{c}{$\text{Hawkes}$ } \\\midrule
			\multirow{2}{*}{$K=2$}& 0.5722 & 0.7167 & 0.7493 & 0.9557 & \textbf{0.9801} & \textbf{0.9776} & 0.5037 & 0.7525 & 0.7741 & \textbf{0.9879} & 0.9710 & 0.9653 \\
 &(0.0387) &(0.0290) &(0.0239) &(0.0164) &(0.0053) &(0.0035) &(0.0548) &(0.0219) &(0.0152) &(0.0110) &(0.0126) &(0.0141) \\\hline
			\multirow{2}{*}{$K=3$}& 0.4117 & 0.5518 & 0.6235 & 0.9047 & 0.9515 & \textbf{0.9660} & 0.3788 & 0.5739 & 0.6405 & \textbf{0.9558} & 0.9316 & 0.9464 \\
 &(0.0447) &(0.0388) &(0.0412) &(0.0179) &(0.0141) &(0.0130)  &(0.0783) &(0.0255) &(0.0239) &(0.0164) &(0.0205) &(0.0184) \\\hline
			\multirow{2}{*}{$K=4$}& 0.2694 & 0.4108 & 0.4556 & 0.8755 & 0.9374 & \textbf{0.9528} & 0.2593 & 0.4290 & 0.4634 & \textbf{0.9256} & 0.9091 & 0.9182 \\
 &(0.0714) &(0.0447) &(0.0436) &(0.0173) &(0.0176) &(0.0170) &(0.0707) &(0.0412) &(0.0311) &(0.0176) &(0.0257) &(0.0155)\\\midrule
			
			$\text{Evaluation}$& \multicolumn{12}{c}{Random Index (RI): the higher the better}\\\midrule
			$\text{Dataset}$ & \multicolumn{6}{c|}{$\text{noHawkes}$ } & \multicolumn{6}{c}{$\text{Hawkes}$ } \\\midrule
			\multirow{2}{*}{$K=2$} & 0.3874 & 0.6114 & 0.6685 & 0.9006 & \textbf{0.9547} & \textbf{0.9517} & 0.3293 & 0.6283 & 0.6832 & \textbf{0.9418} & 0.9274 & 0.9297 \\
 &(0.0917) &(0.0701) &(0.0529) &(0.0184) &(0.0105) &(0.0082)  &(0.0503) &(0.0197) &(0.0173) &(0.0095) &(0.0195) &(0.0130)\\\hline
			\multirow{2}{*}{$K=3$}& 0.2940 & 0.4822 & 0.5265 & 0.8618 & 0.9262 & \textbf{0.9439} & 0.2411 & 0.4995 & 0.5361 & \textbf{0.9251} & 0.8960 & 0.9193\\
 &(0.1342) &(0.0755) &(0.0748) &(0.0296) &(0.0192) &(0.0114)&(0.0640) &(0.0268) &(0.0217) &(0.0087) &(0.0243) &(0.0145)\\\hline
			\multirow{2}{*}{$K=4$}& 0.1182 & 0.3750 & 0.4127 & 0.8043 & 0.8857 & \textbf{0.9128} & 0.0946 & 0.3898 & 0.4189 & \textbf{0.8912} & 0.8626 & 0.8779 \\
 &(0.0424) &(0.0837) &(0.0954) &(0.0247) &(0.0214) &(0.0130) &(0.0775) &(0.0361) &(0.0212) &(0.0143) &(0.0207) &(0.0170)\\\midrule
 
 $\text{Evaluation}$ & \multicolumn{12}{c}{Empirical Intensity Deviation (EID): the lower the better}\\\midrule
 $\text{Dataset}$ & \multicolumn{6}{c|}{$\text{noHawkes}$ } & \multicolumn{6}{c}{$\text{Hawkes}$ } \\\midrule
 			\multirow{2}{*}{$K=2$} & --- & 3.734 & 2.947 & 1.512 & \textbf{0.355} & \textbf{0.358} & --- & 2.787 & 1.963 & \textbf{0.338} & 0.524 & 0.417\\
 & --- &(0.1612) &(0.1789) &(0.0279) &(0.0158) &(0.0126)  & --- &(0.0819) &(0.0768) &(0.0187) &(0.0236) &(0.0241)\\\hline
			\multirow{2}{*}{$K=3$}& --- & 3.960 & 3.255 & 2.173 & 0.484 & \textbf{0.419} & --- & 3.149 & 2.918 & \textbf{0.373} & 0.557 & 0.446 \\
 & --- &(0.1549) &(0.2324) &(0.0557) &(0.0181) &(0.0138)  & --- &(0.1342) &(0.0849) &(0.0235) &(0.0259) &(0.0274)\\\hline
			\multirow{2}{*}{$K=4$} & --- & 3.904 & 3.726 & 2.219 & 0.554 & \textbf{0.472} & --- & 3.321 & 3.135 & \textbf{0.409} & 0.570 & 0.491 \\
 & --- &(0.2449) &(0.2168) &(0.0469) &(0.0202) &(0.0152)  & --- &(0.1817) &(0.1183) &(0.0247) &(0.0518) &(0.0402)\\
			\bottomrule
	\end{tabular}}
	\vspace{2pt}
	\caption{Mean and standard deviation (SD, in bracket) of metrics by clustering (CP, RI) and policy fitting (EID) on synthetic data generated by non-Hawkes and Hawkes process (both up to four intensities). On Hawkes-like data, WGANMM, RLPMM perform no better than Hawkes based model DMMHP; on non-Hawkes data, DMMHP degrades significantly which shows network based models' flexibility. Also network methods show more stability regarding SD.}\label{tab:syn_res}
	\vspace{-14pt}
\end{table}
\subsection{Experiments on Synthetic Data}\label{subsec:syn}
\vspace{-6pt}
We experiment on several synthetic datasets with different $K$ (number of clusters). We generate four kinds of synthetic event sequences in a time interval $[0,T] \quad (T=100)$ using the simulation method in \cite{ogata1981lewis} for TPP. We experiment with different $K$ from $K=2$ to $K=4$, and the results are given by the average of $10$ trails as also shown in Table \ref{tab:syn_res}. The ratio of each cluster size is the same. 
To make a fair comparison with the Hawkes process based models, we experiment on synthetic datasets generated by both non-Hawkes processes and Hawkes processes. 

Specifically, for non-Hawkes dataset, we have sequences generated by a mixture of $K=2$ clusters from $Sine$ intensity and $Negative$-$sine$ intensity, then we add sequences generated by $Constant$ intensity for $K=3$, followed by the cluster from $Bimdoal$ intensity for $K=4$. We list the formulas of the intensity and plot the curve of the ground-truth intensity and learned intensity in supplementary material.

For Hawkes processes, we adopt the conventional Hawkes process as:
$\lambda(t) = \gamma_0 + \alpha\sum_{t\in\tau}e^{-w(t-t')}$.
We also experiment with different $K$ from $K=2$ to $K=4$. In line with \cite{XuNIPS17}, for each trial, the parameters of the intensity function for cluster $k$: $\{\gamma_0^k, \alpha^k\}$ are sampled randomly from $[0,1]$ by keeping $w=1$.In line with \cite{XuNIPS17}, for each trial, the parameters of the intensity function for cluster $k$: $\{\gamma_0^k, \alpha^k\}$ are sampled randomly from $[0,1]$ by keeping $w=1$.

\subsection{Experiments on MemeTracker Data}
\vspace{-6pt}
We collect real-world event sequences from public MemeTracker~\cite{Leskovec:2009:MDN:1557019.1557077} as widely used in TPP works \cite{ZhouAISTATS13,mei2017neural,XiaoNIPS17}. 
It tracks meme diffusion over public media, containing more than 172 million news articles or blog posts. The memes are sentences, such as ideas, proverbs, and the time is recorded when it spreads to certain websites.
We randomly sample 35,000 cascades from MemeTracker to study the diffusion process of the meme since its creation. The memes are supposed generated from different latent policies for discovery. Note one can only use clustering consistency as metrics as there is no ground-truth cluster labels. The results are shown in Table \ref{tab:real_res}.
\begin{table*}[tb!]
	\centering
	\resizebox{0.98\textwidth}{!}{
		\begin{tabular}{c|c|ccccc|ccccc}
			\toprule
			\multicolumn{2}{c|}{Method} & \multicolumn{5}{c|}{WGANMM} & \multicolumn{5}{c}{RLPMM}\\ \midrule
			\multicolumn{2}{c|}{$Cluster\#$}& $K=2$ & $K=3$ & $K=4$ & $K=5$ & $second/iter$ & $K=2$ & $K=3$ & $K=4$ & $K=5$ & $second/iter$ \\\midrule
			\multirow{2}{*}{Dataset} & Synthetic data & $1.8\times 10^4$ & $3.0\times 10^4$ & $5.0\times 10^4$ & --- & $10.32$ sec/iter & $1.0\times 10^4$ & $1.5\times 10^4$ & $2.0\times 10^4$ & --- & $1.125$ sec/iter \\
			& MemeTracker & $0.8\times 10^4$ & $1.2\times 10^4$ & $2\times 10^4$ & $3.0\times 10^4$ & $5.70$ sec/iter & $0.5\times 10^4$ & $0.8\times 10^4$ & $1.5\times 10^4$ & $2.0\times 10^4$ & $0.720$ sec/iter  \\
			\bottomrule
	\end{tabular}}
	\vspace{-4pt}
	\caption{General number of iterations (in order) for training to convergence and time cost per iteration. RLPMM is one order faster. Experiments are conducted on six GeForce GTX 1080 GPUs, for each computing gradients of each cluster in M-step.}\label{tab:runtime}
	\vspace{-15pt}
\end{table*}
\subsection{Findings and Discussion}
We present interpretations to the results as follows.

\textbf{1) Parametric point process vs. neural point process vs. discretized time series} The two neural network based TPP models: RLPMM using reinforcement imitation learning and WGANMM using generative adversarial learning, in general outperform the (implicit) parametric intensity models\footnote{ODE and LS based methods are also called nonparametric TPP as they involve an implicit model to learn intensity forms. In this paper, to simplify the naming, we slightly abuse the term of parametric point process to distinguish them from the neural point process models with significantly more network parameters to learn.}: ODE+GMM, LS+GMM and DMMHP using explicit intensity functions, on both clustering performance and modeling capability. While the performance of time series based method VAR+GMM is the worst. These results show the superiority of the neural model compared with parametric TPP which assume a predefined form with limited model capacity.

On the other hand, as shown in Table \ref{tab:syn_res}, when the data is exactly generated from the predefined point process -- Hawkes processes, the model DMMHP which is based on Hawkes processes model can benefit significantly and even (slightly) outperforms the network based methods including RLPMM and WGANMM. This also suggests the parametric model still have their value when the distribution can be (exactly) known to allow for model specification. The standard deviation in Table \ref{tab:syn_res} also shows the higher stability of neural TPP methods against parametric ones.

\textbf{2) Two-step pipeline vs. joint model} On both synthetic and real-world datasets, joint modeling methods DMMHP, WGANMM and RLPMM outperform two-step VAR/ODE/LS+GMM models which run clustering first followed by learning within each cluster. This shows the utility of an elegant joint learning framework.

\textbf{3) WGANMM vs. RLPMM} It is shown that WGANMM and RLPMM perform relatively close to each other (though often RLPMM outperforms WGANMM) regarding with clustering performance on all metrics, for both synthetic data and real-world dataset. However, we find that the proposed RLPMM shows superior efficiency to WGANMM. In fact, WGANMM adopts the adversarial training framework based on Wasserstein divergence, where both the generator and the discriminator are modeled as dynamic RNN of multiple LSTM cells. In contrast, RLPMM only models the policy as a single LSTM cell, so for the discriminator with an extra Logistic regression layer. As such, RLPMM requires less parameters and computations. Moreover, we find the real-world MemeTracker need notably fewer iterations to converge than synthetic data.

\textbf{4) Learning synthetic data generated by different numbers of policies} Clustering purity, RI and EID all degenerate as the number of policies used for generating the testing data (i.e. intensity functions) grows from $K=2$ to $K=4$ (see the protocol in Section \ref{subsec:syn}). This is no surprise as it becomes more challenging when the sequence set becomes more mixed. While the impact is lessened on the joint modeling methods DMMHP, WGANMM and RLPMM than the two-step models VAR/ODE/LS+GMM. This also holds for the comparison between RLPMM and WGANMM whereby RLPMM need much less additional iterations to converge than WGANMM when cluster number increases from $K=2$ to $K=4$.

\begin{figure}
	\begin{minipage}{0.48\linewidth}
		\raggedright
		\includegraphics[width=\linewidth]{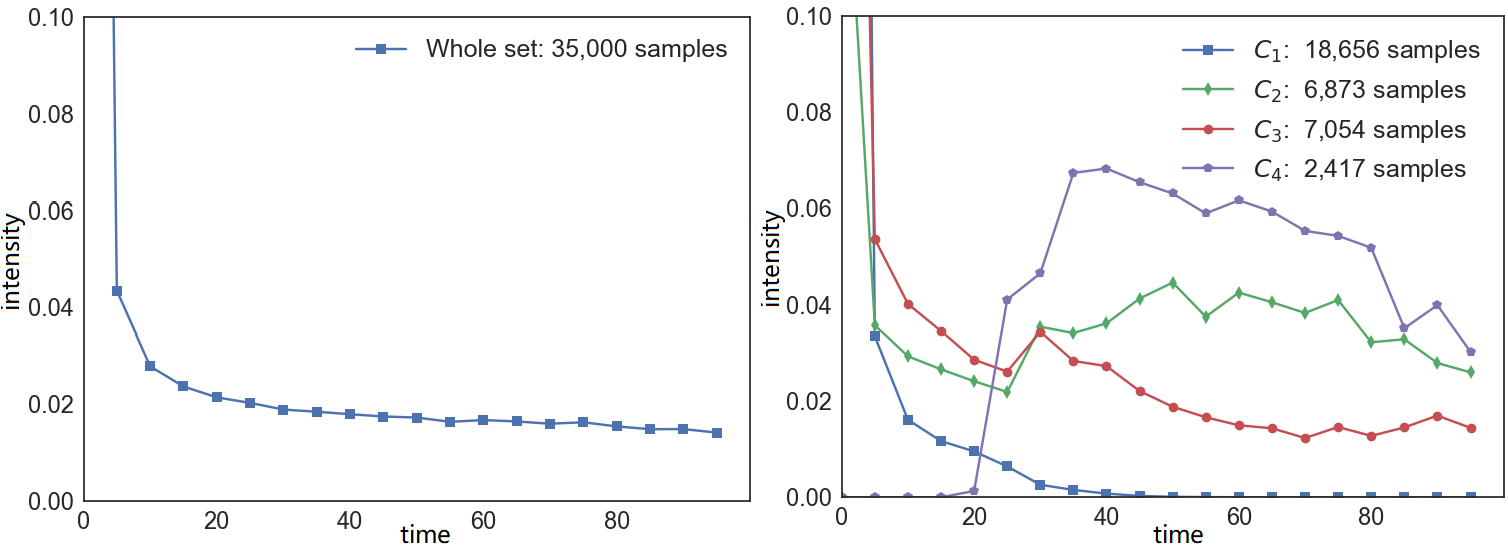}
		\caption{Estimated intensity functions on MemeTracker: a) one intensity for all data; b) $K=4$ intensities by clustering.}\label{fig:reddit}
		\vspace{-18pt}
	\end{minipage}%
	\hfill
	\begin{minipage}{0.49\linewidth}
		\raggedleft
		\resizebox{\textwidth}{!}{
			\begin{tabular}{c|cccccc}
				\toprule
				{Model} & VAR+ & ODE+ & LS+ & \multirow{2}{*}{DMMHP} & \multirow{2}{*}{WGANMM} & \multirow{2}{*}{\textbf{RLMPM}}  \\
				$Cluster\#$& GMM & GMM & GMM &  &   \\ \midrule
				$K=2$ & 0.2194 & 0.4184 & 0.4473 & 0.6190 & 0.8062 & \textbf{0.8169} \\
				$K=3$ & 0.1677 & 0.2491 & 0.3263 & 0.5463 & \textbf{0.7443} & \textbf{0.7415} \\
				$K=4$ & 0.1348 & 0.2116 & 0.2748 & 0.4756 & 0.7024 & \textbf{0.7106} \\
				$K=5$ & 0.0973 & 0.1699 & 0.1925 & 0.4269 & 0.6223 & \textbf{0.6439} \\
				\bottomrule
		\end{tabular}}
		\captionof{table}{Clustering consistency on MemeTracker (in average over trials).}\label{tab:real_res}
		\vspace{-25pt}
	\end{minipage}
\end{figure}

\textbf{4) Learning latent policies of MemeTracker data} Instead of using one less informative TPP model for the whole set of sequences as shown in Fig.~\ref{fig:reddit}(a), we set $K=4$ and plot the empirical intensity functions as discovered by the proposed RLPMM method in Fig.~\ref{fig:reddit}(b). The memes patterns are marked as $C_1$ -- $C_4$, and we show the text statistics of these patterns in supplementary material which reveals a potential for joint modeling with topic model.

For reproducibility, the source code for the proposed RLPMM model and adapted WGANMM model, and the synthetic and MemeTracker dataset is available on Github. \footnote{https://github.com/XXXX/RLPMM}


\vspace{-4pt}
\section{Conclusion and Future Work}
\vspace{-6pt}
Clustering of event sequences in continuous time domain is challenging and has vast application for real-world problems. It is also useful for building event-driven simulators. We study this problem from the reinforcement learning perspective for learning mixture of policies. Our approach involves IRL to learn reward for policy rather than resort to ad-hoc cost for varying-length event sequences.

There are possible extensions in future: i) effective handling with multi-typed event sequences especially noticing the fact that the used generative adversarial imitation learning technique is directly applicable to multi-dimensional TPP. Note the RL method for multi-type TPP in \cite{UpadhyayNIPS18} does not involve IRL technique; ii) joint learning for TPP and topic model for text data associated with events, which is useful in practice such as for MemeTracker; iii) exploring the way of model sharing among clusters for more effective policy learning.

\bibliography{sample-bibliography}

\begin{thebibliography}{10}

\bibitem{ArjovskyWGAN17}
M.~Arjovsky, S.~Chintala, and L.~Bottou.
\newblock Wasserstein generative adversarial nets.
\newblock In {\em ICML}, 2017.

\bibitem{bayer2014learning}
J.~Bayer and C.~Osendorfer.
\newblock Learning stochastic recurrent networks.
\newblock In {\em NIPS Workshop on Advances in Variational Inference}, 2014.

\bibitem{daley2007introduction}
D.~Daley and D.~Vere-Jones.
\newblock {\em An introduction to the theory of point processes: volume II:
  general theory and structure}.
\newblock Springer Science \& Business Media, 2007.

\bibitem{DuKDD16}
N.~Du, H.~Dai, R.~Trivedi, U.~Upadhyay, M.~Gomez-Rodriguez, and L.~Song.
\newblock Recurrent marked temporal point processes: Embedding event history to
  vector.
\newblock In {\em KDD}, 2016.

\bibitem{eichler2017graphical}
M.~Eichler, R.~Dahlhaus, and J.~Dueck.
\newblock Graphical modeling for multivariate hawkes processes with
  nonparametric link functions.
\newblock {\em Journal of Time Series Analysis}, 38(2):225--242, 2017.

\bibitem{ErtekinRPP2015}
S.~Ertekin, C.~Rudin, and T.~H. McCormick.
\newblock Reactive point processes: A new approach to predicting power failures
  in underground electrical systems.
\newblock {\em The Annals of Applied Statistics}, 9(1):122--144, 2015.

\bibitem{farajtabar2017fake}
M.~Farajtabar, J.~Yang, X.~Ye, H.~Xu, R.~Trivedi, E.~Khalil, S.~Li, L.~Song,
  and H.~Zha.
\newblock Fake news mitigation via point process based intervention.
\newblock In {\em ICML}, 2017.

\bibitem{goodfellow2014generative}
I.~Goodfellow, J.~Pouget-Abadie, M.~Mirza, B.~Xu, D.~Warde-Farley, S.~Ozair,
  A.~Courville, and Y.~Bengio.
\newblock Generative adversarial nets.
\newblock In {\em NIPS}, pages 2672--2680, 2014.

\bibitem{han2013transition}
F.~Han and H.~Liu.
\newblock Transition matrix estimation in high dimensional time series.
\newblock In {\em ICML}, pages 172--180, 2013.

\bibitem{hawkes1971point}
A.~G. Hawkes.
\newblock Point spectra of some mutually exciting point processes.
\newblock {\em Journal of the Royal Statistical Society. Series B
  (Methodological)}, 1971.

\bibitem{DBLP:conf/nips/HoE16}
J.~Ho and S.~Ermon.
\newblock Generative adversarial imitation learning.
\newblock In {\em NIPS}, 2016.

\bibitem{Leskovec:2009:MDN:1557019.1557077}
J.~Leskovec, L.~Backstrom, and J.~Kleinberg.
\newblock Meme-tracking and the dynamics of the news cycle.
\newblock In {\em KDD}, 2009.

\bibitem{LewisJNS2011}
E.~Lewis and E.~Mohler.
\newblock A nonparametric em algorithm for multiscale hawkes processes.
\newblock {\em Journal of Nonparametric Statistics}, 2011.

\bibitem{li2014learning}
L.~Li and H.~Zha.
\newblock Learning parametric models for social infectivity in
  multi-dimensional hawkes processes.
\newblock In {\em AAAI}, pages 101--107, 2014.

\bibitem{LiNIPS18}
S.~Li, S.~Xiao, S.~Zhu, N.~Du, Y.~Xie, and L.~Song.
\newblock Learning temporal point processes via reinforcement learning.
\newblock In {\em NIPS}, 2018.

\bibitem{DBLP:journals/pr/Liao05}
T.~W. Liao.
\newblock Clustering of time series data - a survey.
\newblock {\em Pattern Recognition}, 38(11):1857--1874, 2005.

\bibitem{DBLP:journals/classification/Maharaj00}
E.~A. Maharaj.
\newblock Cluster of time series.
\newblock {\em Journal of Classification}, 17(2), 2000.

\bibitem{mei2017neural}
H.~Mei and J.~M. Eisner.
\newblock The neural hawkes process: A neurally self-modulating multivariate
  point process.
\newblock In {\em NIPS}, pages 6757--6767, 2017.

\bibitem{mogren2016c}
O.~Mogren.
\newblock C-rnn-gan: Continuous recurrent neural networks with adversarial
  training.
\newblock {\em arXiv:1611.09904}, 2016.

\bibitem{NgICML00}
A.~Y. Ng and S.~Russell.
\newblock Algorithms for inverse reinforcement learning.
\newblock In {\em ICML}, 2000.

\bibitem{ogata1981lewis}
Y.~Ogata.
\newblock On lewis' simulation method for point processes.
\newblock {\em IEEE Transactions on Information Theory}, 27(1):23--31, 1981.

\bibitem{ozaki1979maximum}
T.~Ozaki.
\newblock Maximum likelihood estimation of hawkes' self-exciting point
  processes.
\newblock {\em Annals of the Institute of Statistical Mathematics},
  31(1):145--155, 1979.

\bibitem{PemantlePS07}
R.~Pemantle.
\newblock A survey of random processes with reinforcement.
\newblock {\em Probability Survey}, 4(0):1--79, 2007.

\bibitem{rand1971objective}
W.~M. Rand.
\newblock Objective criteria for the evaluation of clustering methods.
\newblock {\em Journal of the American Statistical association},
  66(336):846--850, 1971.

\bibitem{DBLP:conf/nips/Rasmussen99}
C.~E. Rasmussen.
\newblock The infinite gaussian mixture model.
\newblock In {\em NIPS}, 1999.

\bibitem{schutze2008introduction}
H.~Sch{\"u}tze, C.~D. Manning, and P.~Raghavan.
\newblock {\em Introduction to information retrieval}, volume~39.
\newblock Cambridge University Press, 2008.

\bibitem{DBLP:journals/corr/TalibiAL17}
A.~Talibi, B.~Achchab, and R.~Lasri.
\newblock Variable selection for clustering with gaussian mixture models: state
  of the art.
\newblock {\em CoRR}, abs/1701.08946, 2017.

\bibitem{Tibshirani2005Cluster}
R.~Tibshirani and G.~Walther.
\newblock Cluster validation by prediction strength.
\newblock {\em Journal of Computational and Graphical Statistics},
  14(3):511--528, 2005.

\bibitem{UpadhyayNIPS18}
U.~Upadhyay, A.~De, and M.~G. Rodriguez.
\newblock Deep reinforcement learning of marked temporal point processes.
\newblock In {\em NIPS}, 2018.

\bibitem{wang2017cascade}
Y.~Wang, H.~Shen, S.~Liu, J.~Gao, and X.~Cheng.
\newblock Cascade dynamics modeling with attention-based recurrent neural
  network.
\newblock In {\em AAAI}, pages 2985--2991, 2017.

\bibitem{WuKDD18}
W.~Wu, J.~Yan, X.~Yang, and H.~Zha.
\newblock Decoupled learning for factorial marked temporal point processes.
\newblock In {\em KDD}, 2018.

\bibitem{XiaoNIPS17}
S.~Xiao, M.~Farajtabar, X.~Ye, J.~Yan, L.~Song, and H.~Zha.
\newblock Wasserstein learning of deep generative point process models.
\newblock In {\em NIPS}, 2017.

\bibitem{XiaoAAAI17}
S.~Xiao, J.~Yan, X.~Yang, H.~Zha, and S.~Chu.
\newblock Modeling the intensity function of point process via recurrent neural
  networks.
\newblock In {\em AAAI}, 2017.

\bibitem{XuNIPS17}
H.~Xu and H.~Zha.
\newblock A dirichlet mixture model of hawkes processes for event sequence
  clustering.
\newblock In {\em NIPS}, 2017.

\bibitem{YanIJCAI18}
J.~Yan, X.~Liu, L.~Shi, C.~Li, and H.~Zha.
\newblock Improving maximum likelihood estimation of temporal point process via
  discriminative and adversarial learning.
\newblock In {\em IJCAI}, 2018.

\bibitem{yan2013towards}
J.~Yan, Y.~Wang, K.~Zhou, J.~Huang, C.~Tian, H.~Zha, and W.~Dong.
\newblock Towards effective prioritizing water pipe replacement and
  rehabilitation.
\newblock In {\em IJCAI}, 2013.

\bibitem{DBLP:conf/ijcai/YuZ18}
Y.~Yu and W.~Zhou.
\newblock Mixture of gans for clustering.
\newblock In {\em IJCAI}, 2018.

\bibitem{ZhouAISTATS13}
K.~Zhou, H.~Zha, and L.~Song.
\newblock Learning social infectivity in sparse low-rank networks using
  multi-dimensional hawkes processes.
\newblock In {\em AISTATS}, 2013.

\bibitem{zhou2013learning}
K.~Zhou, H.~Zha, and L.~Song.
\newblock Learning triggering kernels for multi-dimensional hawkes processes.
\newblock In {\em ICML}, pages 1301--1309, 2013.

\end{thebibliography}
\bibliographystyle{abbrv}

\newpage

\appendix
\section*{Appendix}
\section{Adapting \cite{DBLP:conf/ijcai/YuZ18} to WGANMM for TPP}
For notation clearness, we slightly abuse the notations by assuming the used notations in this subsection is separated to the rest of the paper. Suppose for the $i$-th cluster, training data $D_i=\{x_1,x_2,\dots\}$ is generated by the oracle TPP $r$ and $S_i=\{s_1, s_2, \dots\}$ is generated by the learned TPP $g$, then the Wasserstein distance between the distributions of the two point processes is given by
\begin{equation}\label{eq:w-distance}
W(\mathbb{P}_r,\mathbb{P}_g) = \inf_{\phi\in\Phi(\mathbb{P}_r,\mathbb{P}_g)}\mathbb{E}_{(x,s)\sim\phi}[||x-s||_\star],
\end{equation}
where $\Phi(\mathbb{P}_r,\mathbb{P}_g)$ denotes the set of all joint distributions $\phi(x,s)$ whose marginals are $\mathbb{P}_r$ and $\mathbb{P}_g$, $g$ is learned by minimizing $W(\mathbb{P}_r,\mathbb{P}_g)$.

As Eq.\ref{eq:w-distance} is computationally intractable, hence the dual form is used~\cite{ArjovskyWGAN17} to compute $W(\mathbb{P}_r,\mathbb{P}_g)$ as
\begin{equation}\max_{w\in\mathcal{W},||f_w||_L\leq1} \mathbb{E}_{x\sim\mathbb{P}_r}[f_w(x)] - \mathbb{E}_{s\sim\mathbb{P}_g}[f_w(s)],
\end{equation}
where $f_w$ is the Lipschitz function with parameter $w\in\mathcal{W}$, that assign a value to a event sequence satisfying $1$-Lipschitz constraint $f_w(x)-f_w(s)|\leq||x-s||_\star$ for all $x$ and $s$. As we have supposed that event sequence $s$ is generated by $g_\theta$ with parameter $\theta$ using noisy input $\zeta\sim\mathbb{P}_z$, therefore the objective is to learn a generative model $g_\theta$ by minimize $W(\mathbb{P}_r,\mathbb{P}_g)$ as
\begin{equation}\label{eq:wgan-obj}
\mathop{\min}_\theta \max_{w\in\mathcal{W},||f_w||_L\leq 1}\mathbb{E}_{x\sim\mathbb{P}_r}[f_w(x)] - \mathbb{E}_{\zeta\sim\mathbb{P}_z}[f_w(g_\theta(\zeta))],
\end{equation}
where $f_w$ is \emph{discriminator} and $g_\theta$ is \emph{generator}. Similar to \cite{mogren2016c}, $f_w$ and $g_\theta$ are fulfilled by RNNs.

The \emph{generator} $g_\theta$ transforms a noise input sequence $\zeta=\{z_1,\dots z_n\}$ to generated sequence $s=\{t_1,\dots,t_n\}$ as $g_\theta(\zeta) = s$ using RNN with $n$ LSTM cells:
$$h_i=\phi_g^h(A_g^h z_i + B_g^h h_{i-1} + b_g^h), t_i=\phi_g^t(B_g^t h_i + b_g^t),$$
where $h_i$ is history embedding vector, $\phi_g^h$ and $\phi_g^t$ are activation functions, and parameters $\theta = \{A_g^h,B_g^h, b_g^h, B_g^t,b_g^t\}$. Similarly, the \emph{discriminator} assigns a scalar value $f_w(\rho) = \sum_{i=1}^{n}a_i$ to sequence $\rho=\{t_1,\dots,t_n\}$ ($\rho$ can be real data $x$ or generated one $s$), also by RNN with $n$ cells: $h_i=\phi_f^h(A_f^h z_i + B_f^h h_{i-1} + b_f^h), a_i=\phi_f^t(B_f^t h_i + b_f^t)$, where the parameters of the \emph{discriminator} $w = \{A_f^h,B_f^h, b_f^h, B_f^a,b_f^a\}$.

In general, the adapted WGANMM  and the proposed RLPMM use a similar EM clustering framework. The major differences for our method to WGANMM lie in that WGANMM learns TPP by using Wasserstein distance between event sequences as cost, which may not be optimum for the clustering task at hand. While RLPMM is more like a meta-learning method which adopts IRL -- and the adversarial imitation technique to learn the cost function. Also, RL is used for learning policy of each cluster by RLPMM but WGANMM involves no RL. Moreover, instead of using RNN with multiple cells to learn the discriminators and generators in WGANMM for each cluster, we use RL to learn the latent policy for each cluster. We only need to learn a LSTM cell for the each latent policy, with another LSTM cell to learn the corresponding cost function by IRL. The empirical results in the paper show that our method runs one order faster against WGANMM.

\section{The proof of EM learning convergence}
Due to page limitation, we present the detailed derivation of the EM learning framework and its convergence in this subsection. 

Given a temporal event set $X$ with $M$ observed event sequences: $X=\{x_1,x_2,\dots,x_M\}$ and the  discrete latents i.e. cluster labels $Y=\{y_1, y_2, \dots, y_M\}$ for $y_i\in\{1,2,\dots,N\}$, we suppose that $X$ are generated by a mixture of $N$ experts with a latent policy for each expert, parameterized by $\theta$ as a whole. The log likelihood is:
$$\mathcal{L}(\theta;X,Y) = \log p(X,Y|\theta),$$
where $p(X,Y|\theta)$ is the conditional probability of observing $X,Y$ given parameter $\theta$, and $\theta$ is determined by maximizing the marginal log likelihood of observed $X$:
$$\theta^* = \mathop{\arg\max}\limits_\theta \mathcal{L}(\theta;X)=\mathop{\arg\max}\limits_\theta \log p(X|\theta),$$
where $p(X|\theta)=\int p(X,Y|\theta)dY$ is the marginal probability.

Suppose the latents $Y$ are sampled from an arbitrary valid probability distribution $q(Y)$, then a lower bound $\mathcal{F}(q, \theta)$ of the marginal log likelihood $\mathcal{L}(\theta;X)$ can be obtained by Jensen's inequality:
\begin{align}\label{eq:lowerbound}\notag
\mathcal{L}(\theta;X) = & \log\int p(X,Y|\theta)dY
=  \log \int q(Y)\frac{p(X,Y|\theta)}{q(Y)}dY\\\notag
\geq & \int q(Y)\log\frac{p(Y|X,\theta)p(X|\theta)}{q(Y)}dY\\\notag
= & \int q(Y)\log p(X|\theta)dY + \int q(Y)\log\frac{p(Y|X,\theta)}{q(Y)}dY\\
= & \mathcal{L}(\theta;X) - D_{KL}(q||p),
\end{align}
where $D_{KL}(q||p)$ is the KL-divergence between $q(Y)$ and $p(Y|X,\theta)$, and we have the lower bound:
\begin{equation}
\mathcal{F}(q, \theta)=\mathcal{L}(\theta;X) - D_{KL}(q||p).
\end{equation}
Equation~\ref{eq:lowerbound} gives the relation between hidden variable distribution $q(Y)$ and the likelihood function of the observed data $\mathcal{L}(\theta; X)$: $\forall q(Y):\mathcal{L}(\theta;X)\geq\mathcal{F}(q,\theta)$ and $\mathcal{L}(\theta;X)=\mathcal{F}(q,\theta)$ if and only if $q(Y)=p(Y|X,\theta)$ so that $D_{KL}(q||p)=0$.

Based on Eq.~\ref{eq:lowerbound}, given randomly initialized parameter $\theta^{(0)}$ and arbitrary distribution $q^{(0)}$, we have the following EM procedure that iteratively updates $q^{(k)}$ and parameter $\theta^{(k)}$:

\textbf{E-step}: given model parameter $\theta^{(k)}$, update $q^{(k)}$ to $q^{(k+1)}$ by matching $q$ to posterior $p(Y|X,\theta^{(k)})$, so that $\mathcal{L}(\theta^{(k)})=\mathcal{F}(q^{(k+1)},\theta^{(k)})$

\textbf{M-step}: given $q^{(k+1)}$, update $\theta$: $\theta^{(k+1)} \leftarrow \theta^{(k)} + \nabla \mathcal{F}(q^{(k+1)}, \theta)$ by maximizing $\mathcal{F}(q^{(k+1)}, \theta)$, so that we have: $\mathcal{F}(q^{(k+1)}, \theta^{(k+1)})>\mathcal{F}(q^{(k+1)},\theta^{(k)})$.

By Eq.~\ref{eq:lowerbound}, we have: $\mathcal{L}(\theta^{(k+1)})\geq\mathcal{F}(q^{(k+1)}, \theta^{(k+1)})>\mathcal{F}(q^{(k+1)},\theta^{(k)}) = \mathcal{L}(\theta^{(k)})$, which suggests that for each iteration with E-step and M-step, $\mathcal{L}(\theta;X)$ converges to its maximum. Alternatively perform E-step and M-step iteratively, we can get the optimal solution $\theta^*$ for the model.

\begin{figure}[h]
	\centering
	\subfigure{\includegraphics[width=0.98\textwidth]{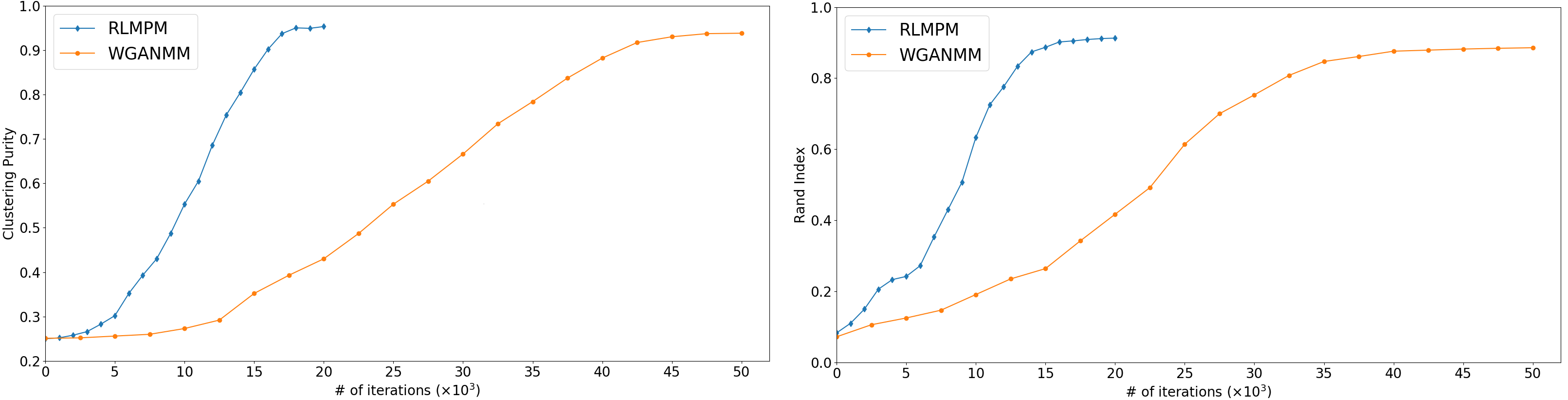}}
	\vspace{-5pt}	
	\caption{Empirical convergence of RLPMM and WGANMM on synthetic data generated by $K=4$ policies. We compute the \emph{clustering purity} and \emph{rand index} every $1\times 10^3$ iterations for RLPMM and $2.5\times 10^3$ iterations for WGANMM. }
	\label{fig:conv}
\end{figure}

For an empirical verification of the convergence of the algorithm, we plot the convergence of the evaluation metric \emph{Clustering Purity} and \emph{Rand Index} on synthetic dataset in Fig.\ref{fig:conv}. As shown in Fig.\ref{fig:conv}, RLPMM model requires less iteration than WGANMM for convergence.

\section{Implementation Trick: Dealing with imbalanced classification}

As we employed the E-step by training a classifier samples generated by the learned policies, the classifier is easy to assign clusters with imbalanced instances, particularly at the beginning of the procedure. The imbalance could get reinforced through the EM procedure. As a result, some policy models receives more and more data and the remaining gets fewer and fewer. Eventually the RLPMM model would converge to an imbalanced solution, 

To fix this issue, at the beginning of the EM procedure, we allow a policy model to explore new by augmenting their training data. After the assignment of clusters $D = \{D_1,\dots, D_N\}$, we augment each cluster data set $D_i$ by adding an amount of instances from $D - D_i$ with the highest posterior probability of belonging to the $i$-th cluster according to the output of the classifier. The amount of the augmented data is reducing along the procedure for the convergence.


\section{Details of Baselines}
For the convenience of reproducibility, we present the technical details of the baselines used in the paper as follows: 

\textbf{1) Gaussian Mixture Model (GMM)}
To tackle the sequential data clustering problem, traditional methods usually implement aggregated time series clustering with discrete time-lagged variable \cite{DBLP:journals/pr/Liao05,DBLP:journals/classification/Maharaj00}. These methods use a probabilistic mixture model to perform sequence clustering with two procedures: firstly extracting features from sequential data, then identifying clusters via GMM \cite{DBLP:journals/corr/TalibiAL17,DBLP:conf/nips/Rasmussen99}. We use three methods to extract features from sequential events for the Gaussian Mixture Model (GMM), including:
\begin{itemize}
	\item \textbf{Vector Auto-Regression (VAR)} The VAR~\cite{han2013transition} model discretizes event sequences to time series, and learns a transition matrix as features for clustering.
	\item \textbf{Ordinary Differential Equation (ODE)} We also use a nonparametric Hawkes process model \cite{LewisJNS2011} based on Ordinary Differential Equation (ODE).
	\item \textbf{Least Squares (LS)} We also test another nonparametric Hawkes model based on contrast function in \cite{eichler2017graphical} relating to the Least Square (LS) problem.
\end{itemize}

Both ODE and LS learn a Hawkes process for each  sequence. In line with the protocol in \cite{XuNIPS17}, we use its parameter $\theta=[\mu, \alpha]$ as feature for each sequence, and employ GMM for clustering.

\textbf{2) Dirichlet Mixture Model of Hawkes Processes (DMMHP)} Learning mixture of policies for TPP has been rarely considered in literature, the most related work to our best knowledge is  DMMHP \cite{XuNIPS17}. It generates event sequences with different clusters from Hawkes processes of different parameters, and uses a Dirichlet distribution as clusters' prior.

Both our model and DMMHP are model-based methods that can accomplish temporal processes clustering. The differences lie in that DMMHP use conventional parameterized Hawkes process and cluster with Latent Dirichlet Allocation (LDA), we propose a novel generative adversarial imitation learning point process model.

\textbf{3) Wasserstein Generative Adversarial Network Mixture Model (WGANMM)} Similar to Gaussian mixture model (GMM), a GAN mixture model in \cite{DBLP:conf/ijcai/YuZ18} has been used for image clustering with fixed sized matrix-like data. In \cite{DBLP:conf/ijcai/YuZ18} $N$ GAN models are trained to capture the each cluster's distribution respectively. To adapt its processing domain from image to event sequences in continuous domain (TPP), we modify vanilla WGANMM by replacing CNN with RNN and introduce the Wassertein distance between point processes as proposed by \cite{XiaoNIPS17} for adversarial learning.

\section{Details of Evaluation Metrics}
The detailed definitions of the metrics used in paper include:

\textbf{1) Clustering Purity (CP)}: Purity is the average of portion of true positive class in each cluster~\cite{schutze2008introduction}:
$$Purity = \frac{1}{M}\sum_{k=1}^{K}\max_{i\in \{1,\dots,K'\}}|\mathcal{W}_k\cap\mathcal{C}_i|,$$
where $\mathcal{W}_k$ is the learned index set belonging to cluster $k$, $\mathcal{C}_i$ is the real index set of sequences belonging to class $i$, $M$ is the total number of sequences. Purity lies in between 0 and 1. Higher purity indicates more concentration in each cluster.

\textbf{2) Rand Index (RI)}: By treating the labels as a clustering ground truth, RI can be used as clustering accuracy (the higher the better), measuring the similarity between the learned sequence clustering and real labels, as given by \cite{rand1971objective}: $RI=\frac{n_{11}+n_{00}}{n}\in[0,1]$, where $n_{11}$ is the number of sequence pairs that are in the same cluster with the same label, and $n_{00}$ is the number of pairs that are in different clusters with different labels.

\textbf{3) Empirical Intensity Deviation (EID)}: To measure the learned latent policy for each cluster, we follow the protocol in \cite{XiaoNIPS17} to compute the deviation of empirical intensity function (accumulated absolute error over time for a windowed period) between the real event sequences and sequences generated by the learned policy, for which the lower the better. The empirical intensity is given by:
$\lambda' (t)=\mathbb{E}(N(t + \delta t)) / \delta t$, where $N(t)$ is the count process for $\lambda(t)$, and the expectation $\mathbb{E}(N(t + \delta t))$ is computed by sufficient number of generated sequences through counting the average number of events during $[t,t+\delta t]$. Note that it can be only applied to synthetic dataset where the ground-truth cluster label is known.

\textbf{4) Clustering Consistency (CC)}: Purity, RI and EID can be used to measure clustering performance when the cluster labels are known as for synthetic sequences. For real-world sequences without labels, we measure the clustering performance by clustering consistency via cross-validation as in~\cite{Tibshirani2005Cluster}.

We test each clustering method with $J=100$ trails. In trial $j$, all sequences are randomly divided into training fold and testing fold. After learning the model from the training fold, we apply the model to the corresponding testing fold. We enumerate all pairs of sequences within a same cluster in the $j$-th trial and count the pairs preserved in all other trials. The clustering consistency is the minimum proportion of preserved pairs over all trials computed by
$$Consistency = \min_{j\in\{1,\dots,J\}}\sum_{j'\neq j}\sum_{(m,m')\in\mathcal{M}_j}\frac{1\{k_m^{j'}=k_{m'}^{j'}\}}{(J-1)|\mathcal{S}_j|},$$
where $\mathcal{S}_j=\{(m,m')|k_m^j = k_{m'}^j\}$ is the sequence pair set within the same cluster in trial $j$, $k_m^j$ is the cluster index of for sequence $m$.

\section{Formulas and plots for non-Hawkes processes}
\begin{figure}[h]
	\centering
	\subfigure{\includegraphics[width=1.0\textwidth]{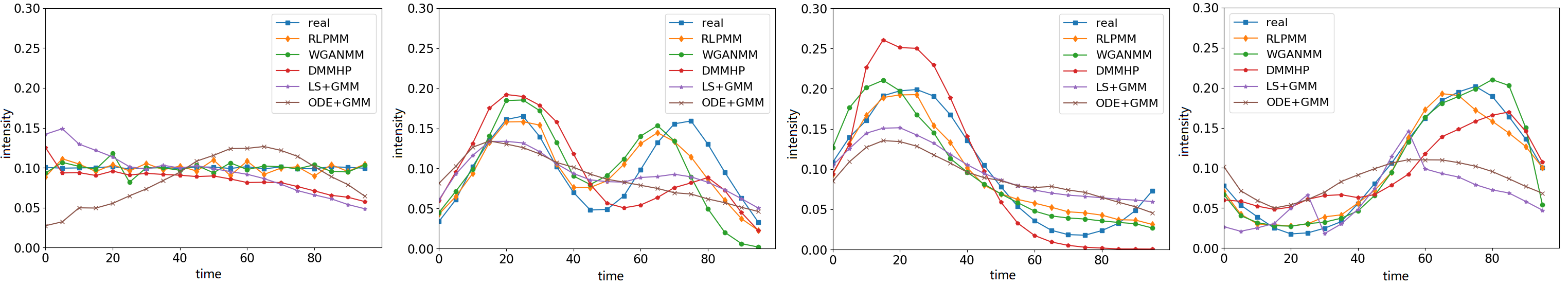}}
	\vspace{-5pt}	
	\caption{Ground truth intensity and estimated ones on synthetic data generated by $K=4$ policies.}
	\label{fig:intensity-synthetic}
\end{figure}

In the experiments with non-Hawkes datasets, the sequences are generated by 4 kinds of intensity functions. The formulas of these functions are: 
\begin{itemize}
	\item \emph{Sine-like}:
	$\lambda(t) = \frac{\sin(\frac{\pi t }{50})}{10} + 0.1, t\in(0,T).$
	\item \emph{Negative-sine-like}:
	$\lambda(t) = - \frac{\sin(\frac{\pi t }{50})}{10} + 0.1, t\in(0,T).$
	\item \emph{Constant}:
	$\lambda(t) = 0.1, t\in(0,T).$
	\item \emph{Bimodal-like}:
	$
	\lambda(t) = \left\{
	\begin{array}{ll}
	0.15\exp(-\frac{t-(\frac{T}{4})^2}{2*(\frac{T}{8})}),&t\in(0,\frac{T}{2}]. \\
	0.15\exp(-\frac{t-(\frac{3T}{4})^2}{2*(\frac{T}{8})}),&t\in(\frac{T}{2},T).\\
	\end{array}
	\right.
	$
\end{itemize}
We also plot the intensity of the ground-truth, and the estimated empirical intensity of RLPMM model and baseline models as shown in Fig.\ref{fig:intensity-synthetic}.  

\section{Interpretation to clustering on MemeTracker}
For the discovered memes patterns marked as $C_1$ -- $C_4$, their statistics of average meme diffusion length and word frequency are listed in Table \ref{tab:nouns}. We make the following interpretations which we believe can at least be partially supported by our results:

\begin{table}[h]
	\centering
	\resizebox{0.98\textwidth}{!}{
		\begin{tabular}{l}
			\toprule
			17.3 \{people, life, God, woman, friend, country, right, America, faith, state, peace, party, earth, future, child\}\\\midrule
			33.9 \{country, people, government, health, economy, campaign, war, market, system, situation, crisis, company, nation, power, market\}  \\\midrule
			77.5 \{people, government, election, McCain, decision, world, challenge, America, policy, security, Europe, law, election, force, Israel\}\\\midrule
			84.1 \{people, world, state, Muslim, threat, industry, terrorist, money, enemy, citizen, Europe, event, America,  safety, Afghanistan\} \\
			\bottomrule
	\end{tabular}}
	\vspace{5pt}
	\caption{Average length (in words) and the top 15 most frequent nouns of memes diffusion cascades, from the discovered four clusters (top to bottom: C1 -- C4) on MemeTracker.}\label{tab:nouns}
	\vspace{-10pt}
\end{table}

i) $C_1$: memes have a wide spread as soon as it is generated, and quickly disappear in around $30$ days. These memes are mostly catchword, e.g. \emph{peace}, \emph{child}, being usually short and clear.

ii) $C_2$: the diffusion intensity decays in $20$ days and then holds in subsequent days. These memes are mostly hot topics like \emph{economy}, \emph{health care}, \emph{job opportunity}, etc..

iii) $C_3$: the diffusion intensity gradually decay for around $40$ days. The memes are mostly long and complete statements and opinions.

iv) $C_4$: the diffusion suspends for short around $20$ days, then begins to diffuse. Though the diffusion processes of $C_3$ and $4$ are different, the average length and top frequent words of $C_4$ are similar to $C_3$ as in Table~\ref{tab:nouns}. It suggests that the diffusion of the long statements and opinions contains two patterns that are quite different from each other.

Such results also show the potential and need for comprehensive modeling by combing RLPMM with topic models on meme content.


\end{document}